# MirrorFlow:
# Exploiting Symmetries in Joint Optical Flow and Occlusion Estimation

Junhwa Hur      Stefan Roth

Department of Computer Science, TU Darmstadt

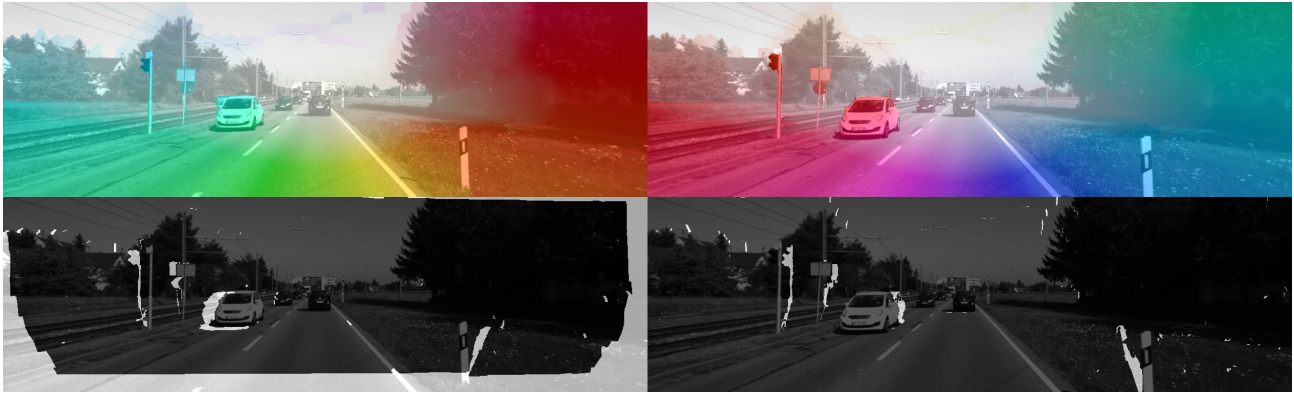

Figure 1. Results of our symmetric optical flow approach given two consecutive images from the KITTI benchmark [12]. Our method jointly predicts forward and backward optical flow *(overlaid on top row, from left to right, color coding of the benchmark)* as well as corresponding occlusion maps for each view *(overlaid on bottom row)*.

## Abstract

*Optical flow estimation is one of the most studied problems in computer vision, yet recent benchmark datasets continue to reveal problem areas of today's approaches. Occlusions have remained one of the key challenges. In this paper, we propose a symmetric optical flow method to address the well-known chicken-and-egg relation between optical flow and occlusions. In contrast to many state-of-the-art methods that consider occlusions as outliers, possibly filtered out during post-processing, we highlight the importance of joint occlusion reasoning in the optimization and show how to utilize occlusion as an important cue for estimating optical flow. The key feature of our model is to fully exploit the symmetry properties that characterize optical flow and occlusions in the two consecutive images. Specifically through utilizing forward-backward consistency and occlusion-disocclusion symmetry in the energy, our model jointly estimates optical flow in both forward and backward direction, as well as consistent occlusion maps in both views. We demonstrate significant performance benefits on standard benchmarks, especially from the occlusion-disocclusion symmetry. On the challenging KITTI dataset we report the most accurate two-frame results to date.*

## 1. Introduction

Optical flow estimation has been studied intensely for several decades. Yet, recent optical flow benchmark datasets [8, 12] reveal challenges that current methods are still struggling to handle. Besides complex large motion and severe illumination changes, occlusions continue to pose a key challenge. Proper occlusion handling, especially in the presence of large motion, is becoming one of the critical factors to a method's success in challenging scenes.

Occlusion estimation is a well-known chicken-and-egg problem that optical flow has been entangled with for a long time [1, 22, 30, 35]. Accurate knowledge of occluded areas is crucial for reliable optical flow estimation in order to prevent non-occluded areas from being adversely affected by occluded pixels. Yet, occlusion is a consequence of motion. Estimating accurate optical flow, conversely, is required for localizing occlusions reliably. We thus posit that their mutual dependency necessitates taking a joint approach and argue that this has not been done to the extent possible.

The majority of recent work instead addresses this challenging joint problem indirectly by considering occlusions as *outliers* of low-level correspondence estimation, *e.g.* [3, 9, 13, 17, 51]. Such approaches aim to mitigate the effects of occlusion by exploiting that occluded pixels gener-ally violate the underlying model assumptions as there are no corresponding pixels in the other frame. Using a robust, truncated penalty in the data term naturally reduces the effects of the high data cost from occlusions, but also more generally from outlier pixels that violate brightness constancy [9, 29, 51]. Checking the forward-backward motion consistency in a subsequent post-processing step and extrapolating flow into inconsistent regions also helps resolving the motion mismatch in the occluded area [9, 11, 17, 29]. These simple procedures have been shown to diminish the influence of occlusions in practice.

Such strategies, however, still cannot completely free optical flow estimation from the ill effects of occlusion. Using a truncated data term leaves the possibility that occluded pixels can be incorrectly matched to other pixels for which the data cost is lower than the truncation constant. Additionally, when extrapolating flow in post-processing, false-positive matches may remain even after the forward-backward consistency check and can cause erroneous estimates to be propagated across a local region [31]. We thus argue that only the accurate localization of occluded regions, formulated as a joint estimation together with the flow, can fundamentally resolve this intertwined problem.

In this paper, we address the chicken-and-egg problem of optical flow and occlusion map estimation and propose a joint energy formulation and optimization method. Our approach directly utilizes their relationship and allows them to leverage each other through estimating forward and backward flow as well as occlusion maps for both directions all together. We exploit two key symmetry properties of the optical flow field and the occlusion map within the two consecutive images: forward-backward flow consistency and occlusion-disocclusion symmetry. These symmetry properties not only couple optical flow with occlusion, but also allow to exploit the geometric and temporal information in the two consecutive images to a greater extent.

The key contributions of our paper are as follows. To the best of our knowledge, we are the first to exploit the occlusion-disocclusion symmetry in joint optical flow and occlusion estimation and show its significant accuracy gains. Second, we demonstrate how this joint, symmetric treatment combined with a piecewise rigid formulation allows estimating optical flow without post-processing. Our experimental results demonstrate state-of-the-art accuracy on public benchmark datasets, where we improve the results especially in occluded areas. For the challenging KITTI dataset we report the most accurate results among two-frame methods to date, outperforming the latest approaches based on high-capacity deep networks [2, 10, 13, 19]. The fact that we are able to do so without employing learning demonstrates the significant benefits of our joint, symmetric flow and occlusion formulation.

## 2. Related Work

We can categorize occlusion handling approaches in optical flow roughly into two families of solutions.

**Occlusions as outliers.** As discussed above, the chicken-and-egg problem of flow and occlusion estimation becomes much simpler if occlusions are treated as outliers that violate the basic optical flow assumptions (*e.g.*, brightness constancy and/or forward-backward flow consistency assumptions). A number of recent algorithms [3, 9, 11, 13, 17, 25, 29] follow a common strategy for outlier filtering. Based on a robust, truncated data term, they *(i)* separately estimate forward and backward flow with an asymmetric method, *(ii)* conduct a bi-directional consistency check, and *(iii)* interpolate flows into the outlier pixels in a post-processing stage, *c.f.* Fig. 2a. With the aid of highly capable interpolation methods [26, 31], this pipeline has been regarded as a well-justified practice. However, occasional failures during post-processing are inevitable and irreversible, and thus constitute a fundamental limitation.

In contrast, our symmetric approach explicitly integrates occlusions into the objective in order to exploit them as an important cue for the flow itself. Also, as shown in Fig. 2b, our integrative approach simultaneously estimates flow in both directions and thus encourages bi-directional consistency of the flow as part of the formulation, which naturally makes any post-processing unnecessary.

**Occlusions in a joint objective.** An outlier is a failure of flow estimation, but an occlusion is a consequence of motion, and can conversely be used as additional evidence for estimating optical flow. Distinguishing between the two opens new possibilities. A number of previous works consider occlusion explicitly in the formulation, but received less attention. We aim to bring them back into focus, revisit their ideas, and highlight the importance of jointly handling occlusions as a feature complementary to other recent trends, including the use of deep learning for appearance matching [2, 10, 13, 19].

All methods in this category begin with including an occlusion variable in their objective. Yet, they differ in the particular characteristics of occlusion utilized and how these are formulated. One basic way to characterize occlusion stems from the observation that *brightness constancy* mostly does not hold in occlusion areas due to the non-existence of corresponding pixels. Several approaches [18, 36, 41, 47] adopt a constant penalty (or truncated cost) in the data term so that it can *(i)* naturally lead to occluded pixels taking the constant penalty rather than a potentially higher matching cost, and *(ii)* explicitly exclude their matching cost from the objective (*e.g.*, as visualized in Fig. (6) of [36]). However, this property alone is not sufficient [47] as it is impossible to discriminate between occluded pixels and pixels with strong illumination changes,

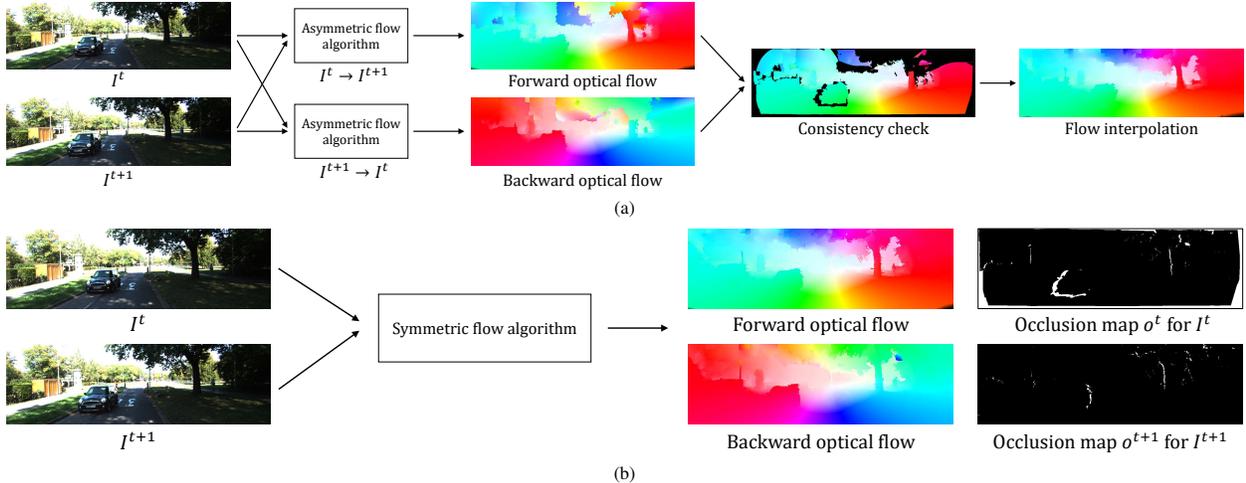

Figure 2. *(a)* A conventional asymmetric approach that requires post-processing. *(b)* Our integrative, symmetric approach.

which both violate the assumption.

Introducing a *forward-backward consistency constraint* into the objective function is another useful strategy. Its benefit is that pixels are forced to be either visible and satisfy the bi-directional flow consistency, or are identified as occlusions [20]. This condition provides an additional cue for joint estimation. Yet, one should not forget that occlusion is a consequence of two different motions causing one pixel to geometrically occlude another. Layered optical flow models [36, 37] or 3D scene flow methods [42] can explicitly model the local depth relationship between layers and estimate occlusions simultaneously. Similarly, one can calculate overlapping areas between two triangular patches and detect occlusions by comparing the photometric cost [21]. Calculating the divergence of the motion field [5] or finding unique configurations of corresponding pixels [22, 41] can be alternative approaches.

Similar to ours, Alvarez *et al.* [1] exploit symmetry properties for jointly estimating optical flow and occlusion, except that their method does not utilize the occlusion and disocclusion symmetry. Interestingly, the most closely related work by Sun *et al.* [38] is from the stereo matching literature, except that forward-backward consistency is not used.

Though all these previous approaches successfully argue the effectiveness of their underlying models, each one omits at least one property that could be utilized. In contrast, our model exploits the relations and symmetry properties between optical flow and occlusion jointly and more completely, leading to significant benefits in accuracy.

## 3. Joint, Symmetric Approach

From two consecutive images, we jointly estimate optical flow maps in both directions and corresponding occlusion maps by fully exploiting their symmetries. The symmetry properties *(i)* couple the two different problem domains, *(ii)* better utilize the available image evidence, and *(iii)* lead to a well-balanced solution during optimization.

The first symmetry we consider is bi-directional motion consistency, *i.e.* motions of corresponding pixels that are visible in both views should be the inverse of one another. Unlike most previous work, we integrate this consistency in the energy, which not only leads to better estimates of both forward and backward flow through iterative optimization, but also obviates conventional post-processing.

Occlusion-disocclusion symmetry is the second property we consider, which geometrically explains how occlusions arise from differing motions of dynamic entities. As illustrated in the third and fourth column of Fig. 3, occlusions and disocclusions demonstrate a symmetry relationship, *i.e.* occlusions in the forward direction correspond to disocclusions in the backward direction and vice versa.

### 3.1. Piecewise rigid optical flow model

Our optical flow model is based on a piecewise rigid representation, which has recently been found to allow the effective regularization of 8-DoF or 9-DoF rigid motion of entities in images, yet still represent both diverse and general motions [15, 18, 28, 44, 51]. We first decompose an image into a set of superpixels [53] as shown in Fig. 3, and estimate the 8-DoF homography motion $\mathbf{H}$ of each superpixel. Each superpixel represents a possible surface in the scene, and the homography represents a locally rigid motion of the surface. Using this model also facilitates formulating the occlusion-disocclusion symmetry property in a comprehensive way, which will be explained below.

### 3.2. Joint energy with symmetries

Given the two consecutive images $I^t$ and $I^{t+1}$ with their superpixel representations, our model jointly estimates *(i)*

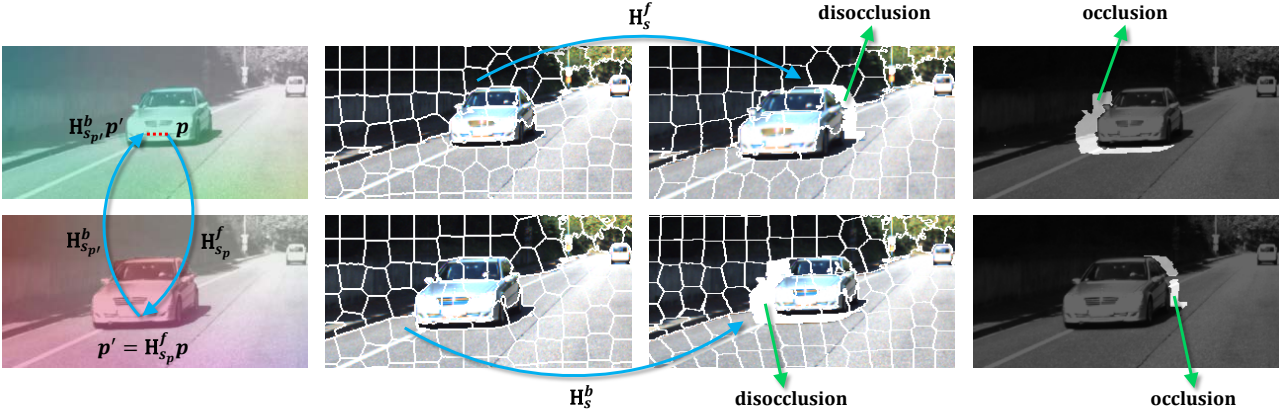

Figure 3. Conceptual explanation of our approach: *(top to bottom)* first frame $I^t$, second frame $I^{t+1}$; *(left to right)* forward/backward flow maps overlaid on each source image, raw superpixel images, warping results given homography motions **H**, and occlusion maps.

the forward motion $\mathbf{H}^f$ (*i.e.*, $I^t \to I^{t+1}$) and the backward motion $\mathbf{H}^b$ (*i.e.*, $I^{t+1} \to I^t$) of each superpixel, and *(ii)* the per-pixel occlusion maps $o^t$ and $o^{t+1}$ for each view. We formulate this through the energy

$$\begin{aligned}
E(\mathbf{H}^f, \mathbf{H}^b, o^t, o^{t+1}) = &\; E_\mathrm{D}(\mathbf{H}^f, \mathbf{H}^b, o^t, o^{t+1}) \\
&+ \lambda_\mathrm{P} E_\mathrm{P}(\mathbf{H}^f, \mathbf{H}^b, o^t, o^{t+1}) \\
&+ \lambda_\mathrm{C} E_\mathrm{C}(\mathbf{H}^f, \mathbf{H}^b, o^t, o^{t+1}) \\
&+ \lambda_\mathrm{S} E_\mathrm{S}(o^t, o^{t+1}),
\end{aligned} \quad (1)$$

which consists of a data term $E_\mathrm{D}$, a pairwise term $E_\mathrm{P}$, a forward-backward consistency term $E_\mathrm{C}$, and an occlusion-disocclusion symmetry term $E_\mathrm{S}$.

### 3.2.1 Data term

The data term accumulates photometric differences across all pixels in both views given the 8-DoF homography motions of superpixels **H** and the per-pixel occlusion masks $o_\mathbf{p}$ in both views:

$$E_\mathrm{D}(\mathbf{H}^f, \mathbf{H}^b, o^t, o^{t+1}) = \sum_{\mathbf{p} \in I^t} D_\mathbf{p}^f + \sum_{\mathbf{p} \in I^{t+1}} D_\mathbf{p}^b \quad (2a)$$

with

$$D_\mathbf{p}^f = \overline{o_\mathbf{p}^t}\, \rho_\mathrm{D}(\mathbf{p}, \mathbf{H}_{s_\mathbf{p}}^f) + o_\mathbf{p}^t \lambda_\mathrm{occ} \quad (2b)$$

$$D_\mathbf{p}^b = \overline{o_\mathbf{p}^{t+1}}\, \rho_\mathrm{D}(\mathbf{p}, \mathbf{H}_{s_\mathbf{p}}^b) + o_\mathbf{p}^{t+1} \lambda_\mathrm{occ}. \quad (2c)$$

For a non-occluded pixel **p** (*i.e.*, $\overline{o_\mathbf{p}} = (1 - o_\mathbf{p}) = 1$), the function $\rho_\mathrm{D}(\mathbf{p}, \mathbf{H}_{s_\mathbf{p}})$ measures the truncated photometric error between pixel **p** and its corresponding pixel $\mathbf{H}_{s_\mathbf{p}}\mathbf{p}$ in the other frame. $\mathbf{H}_{s_\mathbf{p}}$ is the homography motion of superpixel $s_\mathbf{p}$ at pixel position **p**. We use a weighted sum of a gradient constancy term and a ternary transform [34], which is known to be robust under illumination changes [14, 43, 44]. When the corresponding location $\mathbf{H}_{s_\mathbf{p}}\mathbf{p}$ falls outside the image boundary, $\rho_\mathrm{D}(\cdot, \cdot)$ outputs the truncation constant $\tau_\mathrm{D}$.

More specifically, we use a continuous version of the ternary transform, which can improve localization [43] compared to the conventional discrete setting. Furthermore, when calculating the ternary value in the other frame, we transform $7 \times 7$ patches from the reference frame to the other using the given homography $\mathbf{H}_{s_\mathbf{p}}$, and calculate the ternary transform based on the transformed patches. Using this strategy yields a more comprehensive data cost that is invariant to local shape deformation caused by the motion. We observe this to increase the flow accuracy; see supplementary material for details and a quantitative analysis.

For occluded pixels (*i.e.*, $o_\mathbf{p} = 1$), the constant penalty $\lambda_\mathrm{occ}$ is applied so that we can avoid trivial cases in which all pixels are occluded or move outside the image boundary. We set $\lambda_\mathrm{occ} < \tau_\mathrm{D}$ so that pixels whose corresponding location is outside of the image boundary can be naturally inferred as occluded pixels during the optimization.

### 3.2.2 Pairwise term

The pairwise term penalizes the motion differences and occlusion status differences in an 8-neighborhood $N(\mathbf{p})$:

$$E_\mathrm{P}(\mathbf{H}^f, \mathbf{H}^b, o^t, o^{t+1}) = \sum_{\mathbf{p} \in I^t} P_\mathbf{p}^f + \sum_{\mathbf{p} \in I^{t+1}} P_\mathbf{p}^b \quad (3a)$$

with

$$P_\mathbf{p}^f = \sum_{\mathbf{q} \in N(\mathbf{p})} \left( \phi(\mathbf{H}_{s_\mathbf{p}}^f, \mathbf{H}_{s_\mathbf{q}}^f, \bar{\mathbf{p}}) + \lambda_\mathrm{O} \left[ o_\mathbf{p}^t \neq o_\mathbf{q}^t \right] \right) \quad (3b)$$

$$P_\mathbf{p}^b = \sum_{\mathbf{q} \in N(\mathbf{p})} \left( \phi(\mathbf{H}_{s_\mathbf{p}}^b, \mathbf{H}_{s_\mathbf{q}}^b, \bar{\mathbf{p}}) + \lambda_\mathrm{O} \left[ o_\mathbf{p}^{t+1} \neq o_\mathbf{q}^{t+1} \right] \right), \quad (3c)$$

where $\lambda_\mathrm{O}$ is a weight for the pairwise occlusion cost and $[\cdot]$ denotes the Iverson bracket. $\phi(\mathbf{H}_{s_\mathbf{p}}, \mathbf{H}_{s_\mathbf{q}}, \bar{\mathbf{p}})$ measures the difference of two motions induced by the homographies $\mathbf{H}_{s_\mathbf{p}}$ and $\mathbf{H}_{s_\mathbf{q}}$ at neighboring pixels $\mathbf{p}$ and $\mathbf{q}$, evaluated in the middle between the pixels $\bar{\mathbf{p}} = \frac{\mathbf{p}+\mathbf{q}}{2}$. This function distinguishes three different types of pairwise relationships [18, 49, 50] in order to effectively express geometric relations that two neighboring superpixels can have (*e.g.*, coplanar, hinge, others) in terms of their homography motion:

$$\phi(\mathbf{H}_{s_\mathbf{p}}, \mathbf{H}_{s_\mathbf{q}}, \bar{\mathbf{p}}) = w_{\mathbf{p},\mathbf{q}} \cdot \min(\phi_{\text{co}}, \phi_{\text{h}}, \tau_{\text{P}}) \quad (4a)$$

with

$$\phi_{\text{co}} = \frac{1}{|s_\mathbf{p} \cup s_\mathbf{q}|} \sum_{\mathbf{p}_i \in s_\mathbf{p} \cup s_\mathbf{q}} \|\mathbf{H}_{s_\mathbf{p}} \mathbf{p}_i - \mathbf{H}_{s_\mathbf{q}} \mathbf{p}_i\| \quad (4b)$$

$$\phi_{\text{h}} = \|\mathbf{H}_{s_\mathbf{p}} \bar{\mathbf{p}} - \mathbf{H}_{s_\mathbf{q}} \bar{\mathbf{p}}\| + \lambda_{\text{h}}, \quad (4c)$$

where $\tau_{\text{P}}$ is a truncation constant and $\lambda_{\text{h}}$ a constant bias. The intensity-adaptive weight $w_{\mathbf{p},\mathbf{q}} = \exp(-|I(\mathbf{p})-I(\mathbf{q})|/\sigma_w)$ controls the strength of the pairwise term depending on the intensity difference between two neighboring pixels.

The co-planar potential $\phi_{\text{co}}$ calculates the average difference of the homography motions for all pixels within the union of the two superpixels, as they are on the same plane by assumption. The hinge potential $\phi_{\text{h}}$ penalizes the motion difference only for the middle pixel and applies a constant bias $\lambda_{\text{h}}$. For handling the remaining cases such as occlusion or disocclusion, the truncation constant $\tau_{\text{P}}$ is used. Note that the pairwise cost only becomes effective between neighboring pixels that belong to two different superpixels. The cost between neighboring pixels in the same superpixel is naturally zero, because the two pixels have the same associated homography motion.

The pairwise occlusion term simply encourages the spatial smoothness of the occlusion states by penalizing their differences between neighboring pixels. If neighboring pixels have differing occlusion states, the term incurs the constant penalty $\lambda_{\text{O}}$.

### 3.2.3 Forward-backward consistency term

Unlike conventional optical flow algorithms, our model explicitly integrates forward-backward consistency into the objective function [20]:

$$E_{\text{C}}(\mathbf{H}^f, \mathbf{H}^b, o^t, o^{t+1}) = \sum_{\mathbf{p} \in I^t} C_\mathbf{p}^f + \sum_{\mathbf{p} \in I^{t+1}} C_\mathbf{p}^b \quad (5a)$$

with

$$C_\mathbf{p}^f = \overline{o_\mathbf{p}^t}\, \overline{o_{\mathbf{p}'}^{t+1}} \rho_{\text{C}}(\|\mathbf{p} - \mathbf{H}_{s_{\mathbf{p}'}}^b \mathbf{H}_{s_\mathbf{p}}^f \mathbf{p}\|) \quad (5b)$$

$$C_\mathbf{p}^b = \overline{o_\mathbf{p}^{t+1}}\, \overline{o_{\mathbf{p}''}^{t}} \rho_{\text{C}}(\|\mathbf{p} - \mathbf{H}_{s_{\mathbf{p}''}}^f \mathbf{H}_{s_\mathbf{p}}^b \mathbf{p}\|), \quad (5c)$$

where $\mathbf{p}' = \mathbf{H}_{s_\mathbf{p}}^f \mathbf{p}$ and $\mathbf{p}'' = \mathbf{H}_{s_\mathbf{p}}^b \mathbf{p}$. The consistency term penalizes the Euclidean distance between the position of pixel $\mathbf{p}$ and the back-projected position of the corresponding pixel in the other frame, as illustrated in the first column of Fig. 3. Thus, the term encourages the homography matrices from the two corresponding points to be under an inverse relationship, providing a soft constraint on motions in the opposite view, which improves the estimation as shown below. The function $\rho_{\text{C}}(\cdot)$ truncates its input at $\tau_C$ to be robust to possible outliers of flow or occlusion.

The term takes into account all pixels in both views, but applies the penalty only if the pixel $\mathbf{p}$ and its corresponding pixel $\mathbf{p}'$ or $\mathbf{p}''$ in the respective other frame are not occluded. This condition enforces pixels to be either occluded or their motion to satisfy the bi-directional symmetry property. The condition may lead to a trivial solution in which all pixels are marked as becoming occluded such that no penalties arise from this term. However, the constant penalty $\lambda_{\text{occ}}$ in the data term, *c.f*. Eq. (2b) and Eq. (2c), prevents the solution from falling into this trivial case and balances between visible pixels and occlusions.

### 3.2.4 Occlusion-disocclusion symmetry term

The occlusion-disocclusion symmetry term plays the most important role in our model, and allows flow and occlusions to mutually leverage one another. Specifically, the term penalizes cases in which the occlusion-disocclusion symmetry relationship does not hold in both views:

$$E_{\text{S}}(o^t, o^{t+1}) = \sum_{\mathbf{p} \in I^t} o_\mathbf{p}^t \odot N_\mathbf{p}^t + \sum_{\mathbf{p} \in I^{t+1}} o_\mathbf{p}^{t+1} \odot N_\mathbf{p}^{t+1} \quad (6a)$$

with

$$N_\mathbf{p}^t = \left|\left\{\mathbf{p} \mid \mathbf{p} = \mathbf{H}_{s_{\mathbf{p}'}}^b \mathbf{p}', \forall \mathbf{p}' \in I^{t+1}\right\}\right| \quad (6b)$$

$$N_\mathbf{p}^{t+1} = \left|\left\{\mathbf{p} \mid \mathbf{p} = \mathbf{H}_{s_{\mathbf{p}'}}^f \mathbf{p}', \forall \mathbf{p}' \in I^t\right\}\right|. \quad (6c)$$

The XNOR operation (*e.g.*, $0 \odot 0 = 1$ and $1 \odot 0 = 0$) ensures that if a pixel $\mathbf{p}$ is being occluded in one frame, then there cannot be any pixels in the other frame whose motion maps to $\mathbf{p}$. We explicitly detect disocclusion in each view, representing it with the variables $N_\mathbf{p}^t$ and $N_\mathbf{p}^{t+1}$, respectively. $N_\mathbf{p}^t$ denotes the number of pixels that are mapped to $\mathbf{p}$ in $I^t$ when warping pixels in $I^{t+1}$ to $I^t$ given their corresponding motion $\mathbf{H}_s^b$. $N_\mathbf{p}^{t+1}$ is defined analogously by warping pixels from $I^t$ to $I^{t+1}$ given the set of per-superpixel homography motions $\mathbf{H}_s^f$. If no pixel is mapped to $\mathbf{p}$ in $I^{t+1}$ (*i.e.*, $N_\mathbf{p}^{t+1} = 0$), the pixel $\mathbf{p}$ is being disoccluded, as visualized in the third column of Fig. 3. By defining disocclusions solely as a result of the motion, their corresponding variables do not need to be optimized directly, but are indirectly determined by the motion $\mathbf{H}_s^b$ and $\mathbf{H}_s^f$.

In previous studies [22, 41], similar concepts have been introduced in different forms by encouraging unique correspondences between pixels visible in both views. Instead,

**Algorithm 1:** Optimization

    initialize optical flow and occlusion maps
    **for** $n = 1$ to max-iteration **do**
        // *estimate forward flow*
        $\mathbf{H}^f = \arg\min_{\tilde{\mathbf{H}}^f} E(\tilde{\mathbf{H}}^f, \mathbf{H}^b, o^t, o^{t+1})$
        // *estimate occlusion map at time $t+1$*
        update $N^{t+1}$
        $o^{t+1} = \arg\min_{\tilde{o}^{t+1}} E(\mathbf{H}^f, \mathbf{H}^b, o^t, \tilde{o}^{t+1})$
        // *estimate backward flow*
        $\mathbf{H}^b = \arg\min_{\tilde{\mathbf{H}}^b} E(\mathbf{H}^f, \tilde{\mathbf{H}}^b, o^t, o^{t+1})$
        // *estimate occlusion map at time $t$*
        update $N^t$
        $o^t = \arg\min_{\tilde{o}^t} E(\mathbf{H}^f, \mathbf{H}^b, \tilde{o}^t, o^{t+1})$
    **end**

**Algorithm 2:** Optimizing $\mathbf{H}^f$

    INPUT: a current solution of $\mathbf{H}^f$
    // *local expansion move*
    **for** *each local region* $R_i$ **do**
        $\{\mathbf{H}^f_s\} \leftarrow$ propagation($\{\mathbf{H}^f_s \mid s \in R_i\}$)
        $\{\mathbf{H}^f_s\} \leftarrow$ randomization($\{\mathbf{H}^f_s \mid s \in R_i\}$)
    **end**
    // *global expansion move*
    $\mathbf{H}^f \leftarrow$ propagation($\mathbf{H}^f$)

we model symmetry between occlusions and disocclusions while allowing multiple pixels to be mapped to a single location (*i.e.*, $N_\mathbf{p}$ may be greater than 1), which is important when objects in the scene change their apparent size. Moreover, we perform spatial regularization of occlusion states, *c.f.* Eqs. (3b) and (3c).

### 3.3. Optimization

We jointly optimize the two different sets of variables – homography motions (*i.e.*, $\mathbf{H}^f$ and $\mathbf{H}^b$) and occlusion maps (*i.e.*, $o^t$ and $o^{t+1}$) – using a block coordinate descent algorithm, which optimizes the variables alternatingly. As described in Algorithm 1, we first estimate the forward flow $\mathbf{H}^f$. Then, we update $N^{t+1}$ from forward flow $\mathbf{H}^f$ and estimate the occlusion map $o^{t+1}$. The remaining variables $\mathbf{H}^b$, $N^t$, and $o^t$ are updated in turn in a similar manner.

#### 3.3.1 Optimizing $\mathbf{H}^f$ and $\mathbf{H}^b$

Due to the difficulty of optimizing the continuous variables (*e.g.*, stemming from the nonlinearity of the data term), we instead solve a discrete multi-label optimization problem that collects a number of candidate homography motions as proposal sets and then chooses the most suitable motion for each superpixel using fusion moves or $\alpha$-expansion with QPBO [24, 32, 39, 44]. For an efficient optimization, we sequentially run expansion moves locally on subgraphs of superpixels (*e.g.*, a set of neighboring 30 superpixels with 70% overlap between each other)[1] and then globally on all the superpixels as described in Algorithm 2. This strategy follows the similar approach in [40], which defines subgraphs on multiple scales and optimizes them sequentially. We found that this combined local and global optimization strategy yields a faster convergence while avoiding local minima, especially for the planar surface representation with homography-parameterized motions as used here.

The local expansion moves on subgraphs of superpixels [40] consist of two steps, *propagation* and *randomization*, which spatially propagate homography motions and locally refine them. The global expansion moves only conduct the *propagation* step that spatially propagates the locally-refined motions into a broader area.

In each *propagation* and *randomization* step in Algorithm 2, we run expansion moves on the set of input superpixels with each collected set of homography motions. The *propagation* and *randomization* step only differ in how they collect the proposal sets.

*propagation*:
- Randomly sampling $n_p$ homography motions from the input set.
- Randomly sampling $n_p$ homography motions from the corresponding superpixels in the opposite view and taking the inverse motion.

*randomization*:
- Randomly sampling $n_p$ homography motions from the input set and adding perturbations.
- Sampling $n_p$ homography motions by randomly sampling $n_r$ point correspondence pairs and re-estimating the corresponding homography motion.

We set $n_p = 6$ and $n_r = 20$ for the local expansion moves and $n_p = 50$ for the global expansion moves.

#### 3.3.2 Optimizing $o^t$ and $o^{t+1}$

Optimizing the occlusion maps is relatively simple. For instance, once $N^t$ is updated from the backward flow $\mathbf{H}^b$, the binary occlusion map $o^t$ can be estimated via graph-cuts [6, 7, 23] while holding all other variables fixed. The pairwise occlusion term in Eqs. (3b) and (3c) is submodular, thus making standard graph-cuts applicable.

## 4. Experiments

We evaluate the accuracy of our algorithm on the KITTI Optical Flow 2015 benchmark [12] and on the MPI Sintel Flow Dataset [8], both qualitatively and quantitatively. Additionally, we analyze the importance of our symmetric approach by turning each term off and evaluating how significantly it affects the accuracy. For faster convergence, we use DiscreteFlow [29] to initialize our estimation as well as to derive proposals. We automatically tune the parameters (weights $\lambda_P, \lambda_C, \lambda_S, \lambda_O$, truncation thresholds $\tau_D, \tau_P, \tau_C$, and biases $\lambda_{occ}, \lambda_h$) using Bayesian optimization [27] on the training portion of each benchmark.

---
[1]See supplementary material for an analysis of this design choice.

| Method | Non-occluded pixels | | | All pixels | | |
|---|---|---|---|---|---|---|
| | Fl-bg | Fl-fg | Fl-all | Fl-bg | Fl-fg | Fl-all |
| **Ours (MirrorFlow)** | 6.24 % | 12.95 % | 7.46 % | 8.93 % | 17.07 % | **10.29 %** |
| FlowNet2 [19] | 7.24 % | **5.60 %** | **6.94 %** | 10.75 % | **8.75 %** | 10.41 % |
| SDF [2] | **5.75 %** | 18.38 % | 8.04 % | **8.61 %** | 23.01 % | 11.01 % |
| MR-Flow [46] | 6.86 % | 17.91 % | 8.86 % | 10.13 % | 22.51 % | 12.19 % |
| DCFlow [48] | 8.04 % | 19.84 % | 10.18 % | 13.10 % | 23.70 % | 14.86 % |
| SOF [33] | 8.11 % | 18.16 % | 9.93 % | 14.63 % | 22.83 % | 15.99 % |
| DiscreteFlow [29] | 9.96 % | 17.03 % | 11.25 % | 21.53 % | 21.76 % | 21.57 % |

Table 1. **KITTI Optical Flow 2015:** Comparison to top-performing optical flow algorithms in the benchmark in terms of percentages of pixels with an incorrect flow estimate (at the default threshold of 3 pixels). The best and the second best results are in bold and underlined, respectively.

| Method | Final pass | | | Clean pass | | |
|---|---|---|---|---|---|---|
| | EPE all | EPE nocc. | EPE occ. | EPE all | EPE nocc. | EPE occ. |
| DCFlow [48] | **5.119** | **2.283** | 28.228 | 3.537 | 1.103 | 23.394 |
| FlowFieldsCNN [4] | 5.363 | 2.303 | 30.313 | 3.778 | 0.996 | 26.469 |
| MR-Flow [46] | 5.376 | 2.818 | **26.235** | **2.527** | **0.954** | **15.365** |
| S2F-IF [52] | 5.417 | 2.549 | 28.795 | 3.500 | 0.988 | 23.986 |
| RicFlow [16] | 5.620 | 2.765 | 28.907 | 3.550 | 1.264 | 22.220 |
| GlobalPatchCollider [45] | 6.040 | 2.938 | 31.309 | 4.134 | 1.432 | 26.179 |
| **Ours (MirrorFlow)** | 6.071 | 3.186 | 29.567 | 3.316 | 1.338 | 19.470 |
| DiscreteFlow [29] | 6.077 | 2.937 | 31.685 | 3.567 | 1.108 | 23.626 |

Table 2. **MPI Sintel Flow Dataset:** Accuracy in terms of the average end-point error (EPE). Leading algorithms on Final or Clean. Our model performs the second best in the Clean pass.

### 4.1. KITTI Optical Flow 2015

On the KITTI Optical Flow 2015 benchmark, our MirrorFlow algorithm outperforms all two-frame optical flow methods at the time of writing, demonstrating the lowest percentage of flow outliers (*Fl-all*). Table 1 gives detailed numbers, Fig. 4 shows qualitative results. Leveraging our symmetry terms and motion representation based on piecewise homographies, our method also demonstrates the second best results for handling flow on dynamic foreground objects (*Fl-fg*) and handling background motion (*Fl-bg*).

For evaluating the accuracy especially in occluded areas, we cannot directly rely on the KITTI accuracy indicators, as the number of occluded pixels and non-occluded pixels in the testing dataset are unknown. However, by looking at the gap of outlier percentages between all pixels and non-occluded pixels, we can infer that our symmetric method is among the most accurate (probably the most accurate) method in occluded areas (see supplementary material for a more detailed discussion).

One important remark is that our method does not use any learned feature descriptors or semantic information unlike other top-performing algorithms [2, 19, 33, 46, 48]. Even without them, our method significantly outperforms the baseline method [29], which is the next best method not relying on learned descriptors or semantics, and is also used for initialization. We significantly reduce the number of incorrect pixels by more than 50% by virtue of our joint, symmetric formulation. We believe that our method still has room for substantial further improvement by exploiting semantic information or using learned feature descriptors; we leave this for future work.

### 4.2. MPI Sintel Flow Dataset

While we focus on the KITTI dataset with its challenging scenes in the context of autonomous driving, we also evaluate our approach on the MPI Sintel Flow Dataset [8], where approaches based on piecewise rigidity are known to be somewhat disadvantaged. Nevertheless, our method still performs rather competitively, achieving the second place in the Clean pass and the 13[th] place in the Final pass, both at the time of writing. Table 2 gives detailed accuracy numbers. Fig. 5 shows qualitative results using the color code of the Sintel dataset, where we observe rather accurate estimates of occluded regions. The main reason why our method is not as accurate as on the KITTI benchmark is that our planar-rigid motion assumption is not as appropriate for the Sintel Dataset, where the majority of motions are non-rigid and most surfaces are non-planar. Despite of this limitation, our method demonstrates leading results especially on occluded pixels in the Clean pass, which once again confirms the key benefits of our joint flow and occlusion estimation pipeline. Note that while we do not consider this here, the key ideas behind our joint, symmetric approach are not limited to piecewise rigid representations.

### 4.3. Importance of symmetries

We conduct an ablation study to emphasize the contribution of our symmetric formulation and to demonstrate how much each symmetry property contributes to the flow estimation accuracy. As a baseline, we first consider an asymmetric version of our model, which only relies on the data term and the pairwise term (Asymm). Then, we extend it to the symmetric case by estimating flow bi-directionally through enabling the forward-backward consistency term (Symm+c) and the occlusion-disocclusion symmetry term (Symm+s) separately. The full model has both these terms enabled (Symm+cs). We conduct this ablation study on the KITTI Optical Flow 2015 training set.

As shown in Table 3, the occlusion-disocclusion symmetry term has the most significant contribution to the accuracy of flow estimation. The error decreases substantially by about 20%. This observation follows one of our motivations that the accurate localization of occluded areas contributes to estimating flow more accurately. The forward-backward consistency term boosts the quality of the results further, but only when the symmetry term is turned on. This is because the forward-backward consistency relies on accurate estimates of the occlusion regions, which are only available when the occlusion-disocclusion symmetry is considered as well. When both terms are active we achieve the best accuracy, which highlights the benefit of our full symmetric pipeline.

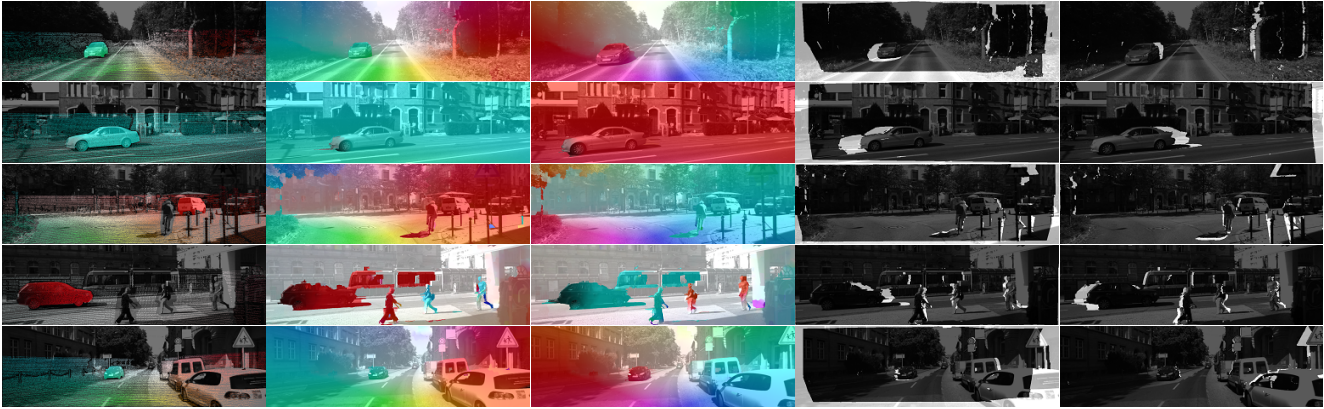

Figure 4. **Qualitative results on KITTI 2015**: *(left to right)* ground truth flow, forward flow, backward flow, occlusion map overlayed on the current frame and the next frame, respectively. Note that the ground truth flow map on KITTI is sparse, and some objects are masked.

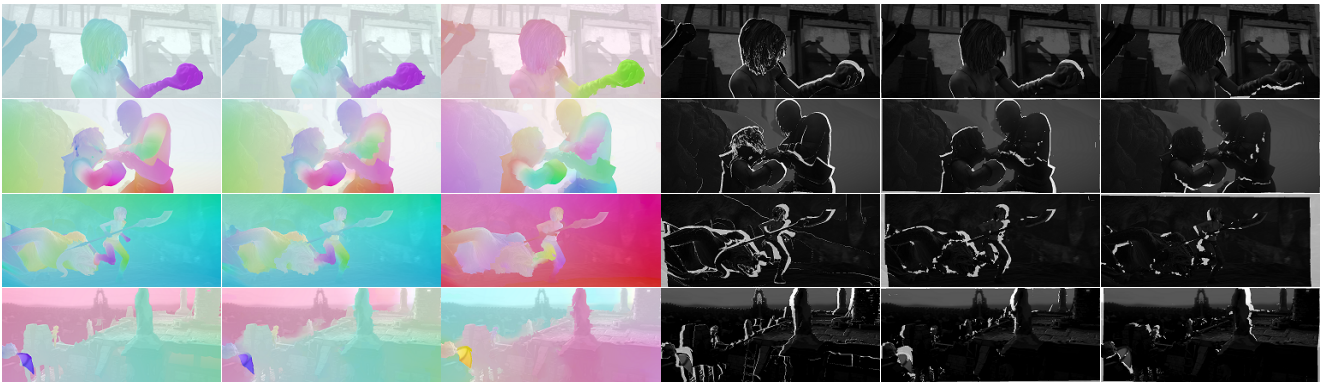

Figure 5. **Qualitative results on Sintel**: *(left to right)* ground truth flow, forward flow, backward flow, ground truth occlusion, occlusion maps, overlayed on the current frame and the next frame, respectively. Note the high agreement between true and estimated occlusions.

|  | Non-occluded pixels | | | All pixels | | |
|---|---|---|---|---|---|---|
| Method | Fl-bg | Fl-fg | Fl-all | Fl-bg | Fl-fg | Fl-all |
| **Symm+cs** | **6.52 %** | **11.72 %** | **7.41 %** | **9.26 %** | **13.94 %** | **9.98 %** |
| Symm+s | 6.73 % | 11.90 % | 7.62 % | 9.49 % | 14.04 % | 10.19 % |
| Symm+c | 10.41 % | 18.17 % | 11.74 % | 13.72 % | 20.40 % | 14.74 % |
| Asymm | 8.39 % | 14.97 % | 9.51 % | 11.82 % | 17.41 % | 12.68 % |

Table 3. **Ablation study for each term** on KITTI 2015 training: The forward-backward consistency term (c) and the occlusion-disocclusion symmetry term (s) both play important roles in our symmetric model. Their combination (cs) shows the best results.

## 5. Conclusion & Future Work

We have proposed a symmetric optical flow method that jointly estimates optical flow in both directions and occlusion maps for each view by exploiting the symmetry properties that they possess in the two consecutive images. We exploit both forward-backward consistency of the flow as well as occlusion-disocclusion symmetry, and formulate a piecewise rigid model. Our results on widely-used public optical flow benchmarks clearly demonstrate that our joint, symmetric approach yields significant improvements in flow estimation accuracy, especially in occluded areas. For the challenging KITTI benchmark, we report leading results even without employing any semantic knowledge or learning of appearance descriptors. We believe that a super-pixel refinement, employing non-rigidity on top of our rigid motion model, or utilizing learned appearance descriptors will lead to further improvements in the future.

**Acknowledgement.** The research leading to these results has received funding from the European Research Council under the European Union's Seventh Framework Programme (FP7/2007–2013) / ERC Grant Agreement No. 307942.

# References

[1] L. Alvarez, R. Deriche, T. Papadopoulo, and J. Sánchez. Symmetrical dense optical flow estimation with occlusions detection. *Int. J. Comput. Vision*, 75(3):371–385, 2007.

[2] M. Bai, W. Luo, K. Kundu, and R. Urtasun. Exploiting semantic information and deep matching for optical flow. In *ECCV*, volume 4, pages 154–170, 2016.

[3] C. Bailer, B. Taetz, and D. Stricker. Flow Fields: Dense correspondence fields for highly accurate large displacement optical flow estimation. In *ICCV*, pages 4015–4023, 2015.

[4] C. Bailer, K. Varanasi, and D. Stricker. CNN-based patch matching for optical flow with thresholded hinge embedding loss. In *CVPR*, 2017.

[5] C. Ballester, L. Garrido, V. Lazcano, and V. Caselles. A TV-L1 optical flow method with occlusion detection. In *DAGM*, pages 31–40, 2012.

[6] Y. Boykov and V. Kolmogorov. An experimental comparison of min-cut/max-flow algorithms for energy minimization in vision. *IEEE T. Pattern Anal. Mach. Intell.*, 26(9):1124–1137, Sept. 2004.

[7] Y. Boykov, O. Veksler, and R. Zabih. Fast approximate energy minimization via graph cuts. *IEEE T. Pattern Anal. Mach. Intell.*, 23(11):1222–1239, Nov. 2001.

[8] D. J. Butler, J. Wulff, G. B. Stanley, and M. J. Black. A naturalistic open source movie for optical flow evaluation. In *ECCV*, volume 4, pages 611–625. 2012.

[9] Q. Chen and V. Koltun. Full Flow: Optical flow estimation by global optimization over regular grids. In *CVPR*, pages 4706–4714, 2016.

[10] A. Dosovitskiy, P. Fischery, E. Ilg, P. Häusser, C. Hazirbas, V. Golkov, P. van der Smagt, D. Cremers, and T. Brox. FlowNet: Learning optical flow with convolutional networks. In *ICCV*, pages 2758–2766, 2015.

[11] D. Gadot and L. Wolf. PatchBatch: A batch augmented loss for optical flow. In *CVPR*, pages 4236–4245, 2016.

[12] A. Geiger, P. Lenz, and R. Urtasun. Are we ready for autonomous driving? The KITTI vision benchmark suite. In *CVPR*, pages 3354–3361, 2012.

[13] F. Güney and A. Geiger. Deep discrete flow. In *ACCV*, volume 1, pages 207–224, 2016.

[14] D. Hafner, O. Demetz, and J. Weickert. Why is the census transform good for robust optic flow computation? In *International Conference on Scale Space and Variational Methods in Computer Vision*, pages 210–221, 2013.

[15] M. Hornáček, F. Besse, J. Kautz, A. Fitzgibbon, and C. Rother. Highly overparameterized optical flow using PatchMatch belief propagation. In *ECCV*, volume 3, pages 220–234. 2014.

[16] Y. Hu, Y. Li, and R. Song. Robust interpolation of correspondences for large displacement optical flow. In *CVPR*, 2017.

[17] Y. Hu, R. Song, and Y. Li. Efficient coarse-to-fine PatchMatch for large displacement optical flow. In *CVPR*, pages 5704–5712, 2016.

[18] J. Hur and S. Roth. Joint optical flow and temporally consistent semantic segmentation. In G. Hua and H. Jégou, editors, *4th Workshop on Computer Vision for Road Scene Understanding and Autonomous Driving*, volume 9913, pages 163–177, 2016. jointly with ECCV 2016.

[19] E. Ilg, N. Mayer, T. Saikia, M. Keuper, A. Dosovitskiy, and T. Brox. FlowNet 2.0: Evolution of optical flow estimation with deep networks. In *CVPR*, 2017.

[20] S. Ince and J. Konrad. Occlusion-aware optical flow estimation. *IEEE T. Image Process.*, 17(8):1443–1451, Oct. 2008.

[21] R. Kennedy and C. J. Taylor. Optical flow with geometric occlusion estimation and fusion of multiple frames. In *EMMCVPR*, pages 364–377, 2015.

[22] V. Kolmogorov and R. Zabih. Computing visual correspondence with occlusions using graph cuts. In *ICCV*, pages 508–515, 2001.

[23] V. Kolmogorov and R. Zabih. What energy functions can be minimized via graph cuts? *IEEE T. Pattern Anal. Mach. Intell.*, 24(2):147–159, Feb. 2004.

[24] V. Lempitsky, S. Roth, and C. Rother. FusionFlow: Discrete-continuous optimization for optical flow estimation. In *CVPR*, 2008.

[25] Y. Li, D. Min, M. S. Brown, M. N. Do, and J. Lu. SPM-BP: Sped-up PatchMatch belief propagation for continuous MRFs. In *ICCV*, pages 4006–4014, 2015.

[26] Y. Li, D. Min, M. N. Do, and J. Lu. Fast guided global interpolation for depth and motion. In *ECCV*, pages 717–733, 2016.

[27] R. Martinez-Cantin. BayesOpt: A Bayesian optimization library for nonlinear optimization, experimental design and bandits. *J. Mach. Learn. Res.*, 15:3735–3739, Jan. 2014.

[28] M. Menze and A. Geiger. Object scene flow for autonomous vehicles. In *CVPR*, pages 3061–3070, 2015.

[29] M. Menze, C. Heipke, and A. Geiger. Discrete optimization for optical flow. In *GCPR*, pages 16–28. 2015.

[30] J.-M. Pérez-Rúa, T. Crivelli, P. Bouthemy, and P. Pérez. Determining occlusions from space and time image reconstructions. In *CVPR*, pages 1382–1391, 2016.

[31] J. Revaud, P. Weinzaepfel, Z. Harchaoui, and C. Schmid. EpicFlow: Edge-preserving interpolation of correspondences for optical flow. In *ICCV*, pages 1164–1172, 2015.

[32] C. Rother, V. Kolmogorov, V. Lempitsky, and M. Szummer. Optimizing binary MRFs via extended roof duality. In *CVPR*, pages 1–8, 2007.

[33] L. Sevilla-Lara, D. Sun, V. Jampani, and M. J. Black. Optical flow with semantic segmentation and localized layers. In *CVPR*, pages 3889–3898, 2016.

[34] F. Stein. Efficient computation of optical flow using the census transform. In *DAGM*, pages 79–86, 2004.

[35] C. Strecha, R. Fransens, and L. Van Gool. A probabilistic approach to large displacement optical flow and occlusion detection. In *Statistical Methods in Video Processing*, pages 71–82. 2004.

[36] D. Sun, C. Liu, and H. Pfister. Local layering for joint motion estimation and occlusion detection. In *CVPR*, pages 1098–1105, 2014.

[37] D. Sun, E. B. Sudderth, and M. J. Black. Layered image motion with explicit occlusions, temporal consistency, and depth ordering. In *NIPS*2010*, pages 2226–2234.

[38] J. Sun, Y. Li, S. B. Kang, and H.-Y. Shum. Symmetric stereo matching for occlusion handling. In *CVPR*, pages 399–406, 2005.

[39] T. Taniai, Y. Matsushita, and T. Naemura. Graph cut based continuous stereo matching using locally shared labels. In *CVPR*, pages 1613–1620, 2014.

[40] T. Taniai, Y. Matsushita, Y. Sato, and T. Naemura. Continuous stereo matching using local expansion moves. *CoRR*, abs/1603.08328, 2016.

[41] M. Unger, M. Werlberger, T. Pock, and H. Bischof. Joint motion estimation and segmentation of complex scenes with label costs and occlusion modeling. In *CVPR*, pages 1878–1885, 2012.

[42] C. Vogel, S. Roth, and K. Schindler. View-consistent 3D scene flow estimation over multiple frames. In *ECCV*, volume 4, pages 263–278, 2014.

[43] C. Vogel, K. Schindler, and S. Roth. An evaluation of data costs for optical flow. In *GCPR*, pages 343–353, 2013.

[44] C. Vogel, K. Schindler, and S. Roth. Piecewise rigid scene flow. In *ICCV*, pages 1377–1384, 2013.

[45] S. Wang, S. R. Fanello, C. Rhemann, S. Izadi, and P. Kohli. The global patch collider. In *CVPR*, pages 127–135, 2016.

[46] J. Wulff, L. Sevilla-Lara, and M. J. Black. Optical flow in mostly rigid scenes. In *CVPR*, 2017.

[47] J. Xiao, H. Cheng, H. Sawhney, C. Rao, and M. Isnardi. Bilateral filtering-based optical flow estimation with occlusion detection. In *ECCV*, pages 211–224, 2006.

[48] J. Xu, R. Ranftl, and V. Koltun. Accurate optical flow via direct cost volume processing. In *CVPR*, 2017.

[49] K. Yamaguchi, D. McAllester, and R. Urtasun. Robust monocular epipolar flow estimation. In *CVPR*, pages 1862–1869, 2013.

[50] K. Yamaguchi, D. McAllester, and R. Urtasun. Efficient joint segmentation, occlusion labeling, stereo and flow estimation. In *ECCV*, volume 5, pages 756–771. 2014.

[51] J. Yang and H. Li. Dense, accurate optical flow estimation with piecewise parametric model. In *CVPR*, pages 1019–1027, 2015.

[52] Y. Yang and S. Soatto. S2F: Slow-to-fast interpolator flow. In *CVPR*, 2017.

[53] J. Yao, M. Boben, S. Fidler, and R. Urtasun. Real-time coarse-to-fine topologically preserving segmentation. In *CVPR*, pages 2947–2955, 2015.

# MirrorFlow:
# Exploiting Symmetries in Joint Optical Flow and Occlusion Estimation
## – Supplementary Material –

Junhwa Hur    Stefan Roth
Department of Computer Science, TU Darmstadt

We here provide additional details on the data term, an analysis of the optimizer, an accuracy analysis in occluded regions, and details on the processing time.

## A. Details on the Data Term

In Eqs. (2b) and (2c) of the main paper, the function $\rho_\text{D}(\mathbf{p}, \mathbf{H}_{s_\mathbf{p}})$ measures the photometric error between a pixel $\mathbf{p}$ and its corresponding pixel $\mathbf{H}_{s_\mathbf{p}}\mathbf{p}$ in the other frame. For example, given a homography motion $\mathbf{H}_{s_\mathbf{p}}^f$, the data cost for pixel $\mathbf{p}$ in $I^t$ is given as

$$\rho_\text{D}^f(\mathbf{p}, \mathbf{H}_{s_\mathbf{p}}^f) = \min\left\{\rho_l\big(\phi(\mathbf{p}, \mathbf{H}_{s_\mathbf{p}}^f)\big), \tau_\text{D}\right\} \quad (7\text{a})$$

with

$$\phi(\mathbf{p}, \mathbf{H}_{s_\mathbf{p}}^f) = \quad (7\text{b})$$
$$\alpha_D \sum_{\mathbf{y} \in \{-3,\ldots,3\}^2} f\bigg(\underbrace{T\big(I^t(\mathbf{p}+\mathbf{y}) - I^t(\mathbf{p})\big)}_{\text{ternary value at } \mathbf{p} \text{ in } I^t}$$
$$- \underbrace{T\big(I^{t+1}(\mathbf{H}_{s_\mathbf{p}}^f(\mathbf{p}+\mathbf{y})) - I^{t+1}(\mathbf{H}_{s_\mathbf{p}}^f\mathbf{p})\big)}_{\text{ternary value at } \mathbf{H}_{s_\mathbf{p}}^f\mathbf{p} \text{ in } I^{t+1}}\bigg)$$
$$+ (1-\alpha_D)\underbrace{|\nabla I^{t+1}(\mathbf{H}_{s_\mathbf{p}}^f\mathbf{p}) - \nabla I^t(\mathbf{p})|}_{\text{gradient constancy penalty}}, \quad (7\text{c})$$

which is the weighted sum of the ternary transform and gradient constancy penalty.

Deviations are penalized by a Lorentzian penalty $\rho_l(x) = \alpha_l \log((1+x^2)/2\sigma_l^2)$, truncated at $\tau_\text{D}$. The idea behind function $\phi(\mathbf{p}, \mathbf{H}_{s_\mathbf{p}}^f)$ is to calculate the Hamming distance of two $7 \times 7$ ternary patches, one around pixel $\mathbf{p}$ in $I^t$ and one around the corresponding pixel $\mathbf{H}_{s_\mathbf{p}}^f\mathbf{p}$ in $I^{t+1}$. Unlike the conventional ternary transform [34], we use a continuous variant. Specifically, we relax the definition of the Hamming distance and adopt the sigmoid function

$$T(x) = \frac{2}{1+\exp(-\sigma_T x)} - 1 = \frac{1-\exp(-\sigma_T x)}{1+\exp(-\sigma_T x)} \quad (8)$$

instead of a true ternary value, and use the Geman-McClure function [54] to score the differences in the ternary signature between the patches:

$$f(x) = \frac{x^2}{(\sigma_f + x^2)}. \quad (9)$$

As shown in Figs. 6a and 6b, these continuous functions approximate the conventional discrete setting, but they assess subtle brightness variations more naturally when their input is near zero. In other words, they are still robust, but less brittle than the original Hamming-based definition.

Furthermore, when calculating the ternary value at point $\mathbf{H}_{s_\mathbf{p}}^f\mathbf{p}$ in the other frame, we calculate it not on the conventional ternary patch that is centered at the transformed point, but on a transformed patch. Eq. (7b) precisely expresses how to calculate the ternary value on the warped patch (*i.e.*, referring the intensity at $\mathbf{H}_{s_\mathbf{p}}^f(\mathbf{p}+\mathbf{y})$ instead of $\mathbf{H}_{s_\mathbf{p}}^f\mathbf{p}+\mathbf{y}$). Similar to a classical iterative-warping scheme, this strategy yields a more comprehensive data cost that is invariant to local shape deformation caused by the motion.

We observe that the two practices above increase the flow accuracy. Table 4 compares the flow accuracy of three different ways of calculating the ternary value: *(i)* our standard implementation including both the continuous ternary variant and the patch transformation (*standard*), *(ii)* the standard implementation with the conventional discrete setting of the ternary transform (*discrete*) but with patch transformation, *(iii)* the standard implementation without the patch

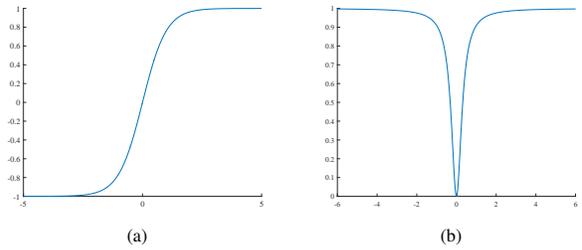

Figure 6. *(a)* Sigmoid function. *(b)* Geman-McClure function.

| | Non-occluded pixels | | | All pixels | | |
|---|---|---|---|---|---|---|
| Method | Fl-bg | Fl-fg | Fl-all | Fl-bg | Fl-fg | Fl-all |
| **standard** | **6.52 %** | **11.72 %** | **7.41 %** | **9.26 %** | **13.94 %** | **9.98 %** |
| discrete | 7.11 % | 12.26 % | 7.99 % | 9.88 % | 14.57 % | 10.60 % |
| w/o transformation | 7.29 % | 12.35 % | 8.16 % | 10.41 % | 14.80 % | 11.08 % |

Table 4. Evaluation of different methods for computing the ternary census on the KITTI training set. See text for details.

transformation (*w/o transformation*) but with the continuous variant.

As presumed, using the continuous ternary variant and the transformed ternary patch both yield better accuracy by reducing the number of flow errors by about 5.85 % and 9.93 % respectively. This experiment has been conducted on the KITTI Optical Flow 2015 training set.

## B. Analysis of the Optimizer

As discussed in Sec. 3.3.1 of the main paper, we first collect a set of proposals when assigning homography motions for each superpixel, and sequentially run expansion moves on each subgraph of superpixels to allow for an efficient optimization. We assemble a set of subgraphs in a way that each subgraph consists of neighboring 30 superpixels with 70 % overlap between each other. We empirically found that this is an advantageous setting in our problem.

Choosing the number of superpixels in each subgraph affects both flow accuracy and energy at convergence. When the number of superpixels is high, proposals can be propagated into broader regions, but the algorithm has a smaller chance of finding locally optimal homography motions, which results in slower convergence. On the other hand, when the number of superpixels is low, locally optimal motions can lower the energy level more quickly, but the optimization can get stuck in local optima as it propagates labels only in small regions, which eventually leads to higher flow error rates.

Figure 7 and Fig. 8 demonstrate the energy and the flow error rate (on KITTI 2015 training), respectively, versus the processing time, depending on the number of superpixels in each subgraph. Each dot on a graph represents an iteration step. These two figures illustrate the trade-off described above. We found that having 30 superpixels for each subgraph yields the lowest flow error rates.

Choosing the size of overlapping regions between subgraphs also incurs a trade-off: When the size is getting bigger, the proposals can be propagated more effectively between subgraphs, which helps finding lower energy solutions in fewer iterations. However, it requires more processing time per iteration because the size of the subgraphs is increased. When the size of overlaps gets smaller, on the other hand, less processing time per iteration is needed, but the optimizer propagates proposals through subgraphs less effectively, leading to more iterations being required.

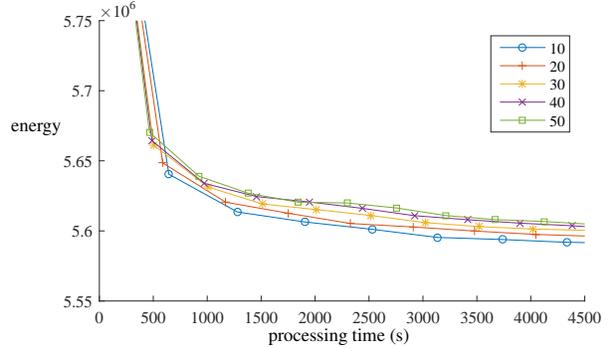

Figure 7. Overall energy depending on the number of superpixels in each subgraph.

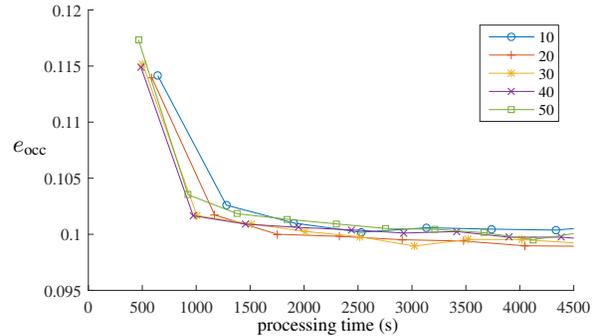

Figure 8. The estimated flow error rates depending on the number of superpixels in each subgraph.

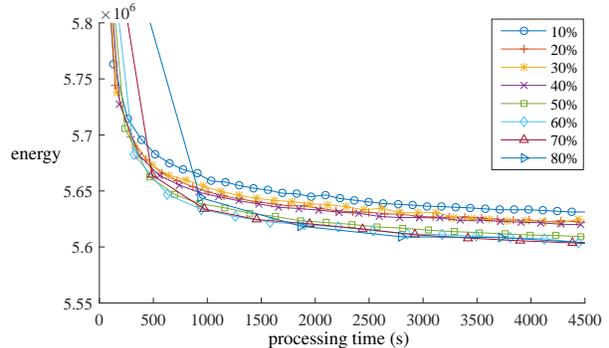

Figure 9. Overall energy depending on the overlap setting.

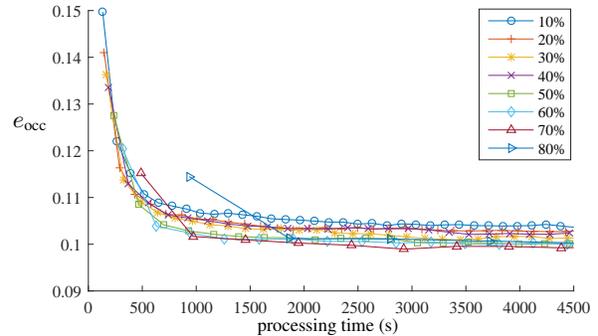

Figure 10. The estimated flow error rates depending on the overlap setting.

Figure 9 and Fig. 10 demonstrate the energy and the flow error rate (on KITTI 2015 training), respectively, versus processing time, depending on the overlap size between subgraphs. As in Fig. 9, if the overlap size is more than 50 %, the energy is converging to a lower value, but with a similar speed. Figure 10 demonstrates that having 70 % of overlap between subgraphs yields the lowest flow error rates in the same processing time. However, please note that the flow accuracy differences between the settings are not very significant ($< 5$ %).

## C. Performance in Occluded Regions

We analyze the flow estimation accuracy of top-performing algorithms including ours especially in occluded regions on the KITTI Optical Flow 2015 benchmark [12]. Unlike the MPI Sintel Flow Dataset [8], the KITTI benchmark does not explicitly provide the statistics for occluded areas. Thus, we indirectly deduce them.

To that end, let us define the variables $n_{\text{all}}, n_{\text{noc}}, n_{\text{occ}}, e_{\text{all}}, e_{\text{noc}}$, and $e_{\text{occ}}$ as follows:

- $n_{\text{all}}$ : no. of all pixels considered in evaluation
- $n_{\text{noc}}$ : no. of non-occluded pixels
- $n_{\text{occ}}$ : no. of occluded pixels
- $e_{\text{all}}$ : no. of all pixels with an incorrect flow estimate
- $e_{\text{noc}}$ : no. of non-occluded pixels with an incorrect flow estimate
- $e_{\text{occ}}$ : no. of occluded pixels with an incorrect flow estimate.

Then, the following equations naturally hold:

$$n_{\text{all}} = n_{\text{noc}} + n_{\text{occ}} \quad (10\text{a})$$
$$e_{\text{all}} = e_{\text{noc}} + e_{\text{occ}}. \quad (10\text{b})$$

We are interested in estimating the flow error rate in occluded areas, $e_{\text{occ}}/n_{\text{occ}}$. From Eqs. (10a) and (10b) we have

$$\frac{e_{\text{occ}}}{n_{\text{occ}}} = \frac{e_{\text{all}} - e_{\text{noc}}}{n_{\text{all}} - n_{\text{noc}}} = \frac{\frac{e_{\text{all}}}{n_{\text{all}}} - \frac{e_{\text{noc}}}{n_{\text{all}}}}{1 - \frac{n_{\text{noc}}}{n_{\text{all}}}} = \frac{\frac{e_{\text{all}}}{n_{\text{all}}} - \frac{e_{\text{noc}}}{n_{\text{noc}}} \frac{n_{\text{noc}}}{n_{\text{all}}}}{1 - \frac{n_{\text{noc}}}{n_{\text{all}}}}. \quad (11)$$

Given that we do not know the ratio of non-occluded pixels, we substitute $n_{\text{noc}}/n_{\text{all}}$ with $\alpha$ in Eq. (11) and obtain

$$\frac{e_{\text{occ}}}{n_{\text{occ}}} = \frac{1}{1-\alpha} \frac{e_{\text{all}}}{n_{\text{all}}} - \frac{\alpha}{1-\alpha} \frac{e_{\text{noc}}}{n_{\text{noc}}}, \quad (12)$$

where the values $e_{\text{all}}/n_{\text{all}}$ and $e_{\text{noc}}/n_{\text{noc}}$ are the flow error rates on all pixels and non-occluded pixels, respectively, which can be found in Table 1 of the main paper. Therefore, we can indirectly infer the flow error rate in occluded areas based on the statistics from Table 1 and in terms of $\alpha = n_{\text{noc}}/n_{\text{all}}$, which denotes the (unknown) ratio of the number of non-occluded pixels to that of all pixels that are considered in the evaluation.

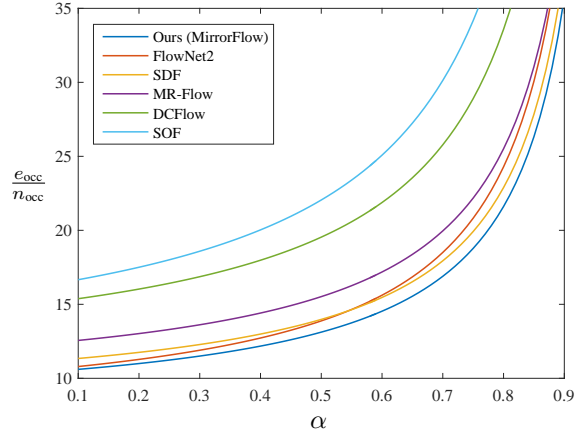

Figure 11. The estimated flow error rates of top-performing algorithms in occluded regions with respect to $\alpha = n_{\text{noc}}/n_{\text{all}}$.

| Method | Fl-all in occluded pixels (estimates) |
|---|---|
| **Ours (MirrorFlow)** | **28.19 %** |
| SDF [2] | 29.80 % |
| FlowNet2 [19] | 32.36 % |
| MR-Flow [46] | 36.23 % |
| DCFlow [48] | 44.47 % |
| SOF [33] | 54.33 % |

Table 5. Estimated flow errors for occluded pixels (when $\alpha = 0.8635$). Our method demonstrates the lowest error among all published two-frame methods on the KITTI benchmark.

In Fig. 11 we now plot the estimated flow error rates of top-performing algorithms in occluded regions by varying the unknown ratio $\alpha$. We observe that our MirrorFlow approach consistently shows the lowest error rates among the top-performing algorithms regardless of values of the ratio $\alpha$. Considering that the ratio $\alpha$ for the KITTI 2015 training set is $\alpha = 0.8635$, we can confidently infer that our method demonstrates the lowest optical flow error among the top-performing algorithms on the KITTI benchmark. Table 5 gives the estimated results assuming the same $\alpha$ as on the training set.

## D. Processing Time

For processing a $1226 \times 370$ image, the algorithm takes around 40 minutes on a single core until the accuracy no longer increases (tested on Intel Xeon CPU E5-2650 2.20GHz). Yet, the algorithm can be easily parallelized because the local subgraphs that do not overlap with each other can be processed at the same time [40]. Using 4 cores, the runtime decreases down to 11 minutes.

The main bottleneck is calculating the ternary transform in the data term. We calculate the ternary census on transformed patches, which needs to be done for every different homography motion. When just using a plain data term (penalizing intensity + gradient differences), the runtime is only 4 minutes. CNN-based learned descriptors also have the potential to lead to a speedup as a future work.

## References

[54] M. J. Black and A. Rangarajan. On the unification of line processes, outlier rejection, and robust statistics with applications in early vision. *Int. J. Comput. Vision*, 19(1):57–91, July 1996.